%% file: iclr2026_conference.tex
\documentclass{article} 
\usepackage{iclr2026_conference,times}

\input{math_commands.tex}

\usepackage{hyperref}
\usepackage{url}
\usepackage{tikz}
\usetikzlibrary{positioning}
\usepackage{booktabs}

\title{Learning Proposes, Geometry Disposes: A Modular Framework for Efficient Spatial Reasoning}

\author{Haichao Zhu \\
\\ Reality Vision
\texttt{hczhu@reality.vision} \\
\And
Zhaorui Yang \\
Department of Computer Science and Engineering \\
University of California, Riverside  \\
\texttt{zyang247@cs.urc.edu} \\
\And
Qian Zhang \\
Department of Computer Science and Engineering \\
University of California, Riverside  \\
\texttt{qzhang@cs.urc.edu} \\
}

\iclrfinalcopy 
\begin{document}

\maketitle

\begin{abstract}
\input{abstract}

\end{abstract}

\section{Introduction}
\input{introduction}

\section{Related Work}
\input{related}

\section{Problem Setup and Design Principles}
\input{problem}

\section{Pipeline Design}
\input{framework}

\section{Experimental Setup}
\input{experiments}

\section{Results}
\input{results}

\section{Discussion}
\input{discussion}

\bibliography{iclr2026_conference}
\bibliographystyle{iclr2026_conference}

\appendix
\input{appendix}

\end{document}

%% file: math_commands.tex
\usepackage{amsmath,amsfonts,bm}

\def\eqref#1{equation~\ref{#1}}

\def\1{\bm{1}}

\DeclareMathAlphabet{\mathsfit}{\encodingdefault}{\sfdefault}{m}{sl}
\SetMathAlphabet{\mathsfit}{bold}{\encodingdefault}{\sfdefault}{bx}{n}



%% file: abstract.tex
Spatial perception aims to estimate camera motion and scene structure from visual observations, a problem traditionally addressed through geometric modeling and physical consistency constraints.
Recent learning-based methods have demonstrated strong representational capacity for geometric perception and are increasingly used to augment classical geometry-centric systems in practice.
However, whether learning components should directly replace geometric estimation or instead serve as intermediate modules within such pipelines remains an open question.

In this work, we address this gap and investigate an end-to-end modular framework for effective spatial reasoning, where learning proposes geometric hypotheses, while geometric algorithms dispose estimation decisions. In particular, we study this principle in the context of relative camera pose estimation on RGB-D sequences. Using VGGT~\cite{wang2025vggt} as a representative learning model, we evaluate learning-based pose and depth proposals under varying motion magnitudes and scene dynamics, followed by a classical point-to-plane RGB-D ICP as the geometric backend. Our experiments on the TUM RGB-D benchmark~\cite{sturm12iros} reveal three consistent findings: (1) learning-based pose proposals alone are unreliable; (2) learning-proposed geometry, when improperly aligned with camera intrinsics, can degrade performance; and (3) when learning-proposed depth is geometrically aligned and followed by a geometric disposal stage, consistent improvements emerge in moderately challenging rigid settings.

These results demonstrate that geometry is not merely a refinement component, but an essential arbiter that validates and absorbs learning-based geometric observations. Our study highlights the importance of modular, geometry-aware system design for robust spatial perception.



%% file: introduction.tex
Geometric reasoning has long been the foundation of visual localization and mapping systems. Classical formulations based on multi-view geometry, camera models, and geometric optimization provide strong inductive biases and physical interpretability, as systematically established in Multiple View Geometry in Computer Vision~\cite{hartley2003multiple}. These principles underpin a wide range of successful visual odometry and SLAM systems, from early monocular pipelines such as MonoSLAM~\cite{davison2007monoslam} and PTAM~\cite{klein2007parallel} to modern feature-based frameworks like ORB-SLAM~\cite{mur2015orb}. In such systems, explicit geometric optimization—e.g., bundle adjustment~\cite{triggs1999bundle} or iterative closest point (ICP)~\cite{besl1992method}—serves as the ultimate authority for enforcing geometric consistency.

More recently, learning-based approaches have demonstrated strong representational capacity for geometric perception, including learned correspondence and optimization frameworks such as DROID-SLAM~\cite{teed2021droid}, as well as learning-based depth and motion estimation methods such as DeepV2D~\cite{teed2018deepv2d} and VGGT~\cite{wang2025vggt}. These advances raise a fundamental question: \textbf{how should learning be integrated into classical geometric pipelines, and what role should geometry continue to play in modern spatial perception systems?}

Despite the success of both classical geometry and recent learning-based methods, the division of responsibility between the two remains ambiguous. A common trend is to treat learning as a direct replacement for geometric pose estimation, predicting camera motion end-to-end from raw images~\cite{wang2021tartanvo,wang2017deepvo}. However, such approaches often lack explicit mechanisms for enforcing geometric consistency, leading to unstable behavior under large baselines, distribution shifts, or dynamic scenes~\cite{teed2018deepv2d,teed2021droid}. In contrast, geometric optimization provides a principled way to validate hypotheses through well-defined geometric constraints—i.e., rigid-body geometry and multi-view consistency—but its effectiveness depends critically on the quality of geometric observations~\cite{besl1992method,triggs1999bundle}.

A natural question is whether learning and geometry should be optimized jointly in an end-to-end manner. While appealing in principle, such joint optimization assumes that learning-based predictions provide reliable gradients and consistent geometric signals. In practice, learning-based geometric predictions may suffer from hallucination or overconfident errors that violate rigid-body assumptions, causing optimization to converge to physically inconsistent solutions or fail altogether. When erroneous learning signals are tightly coupled with optimization, geometry can no longer act as an effective safeguard.

Motivated by this tension, we advocate a simple but explicit system principle: learning should propose geometry, while geometry should dispose pose. Explicitly separating learning-based proposal from geometric decision preserves the ability of geometry to act as a robust arbiter. Learning is allowed to suggest rich geometric hypotheses, but these hypotheses are validated only through explicit geometric consistency. This separation prevents learning-induced hallucinations from dominating the optimization process and ensures that pose estimation remains grounded in interpretable geometric constraints.

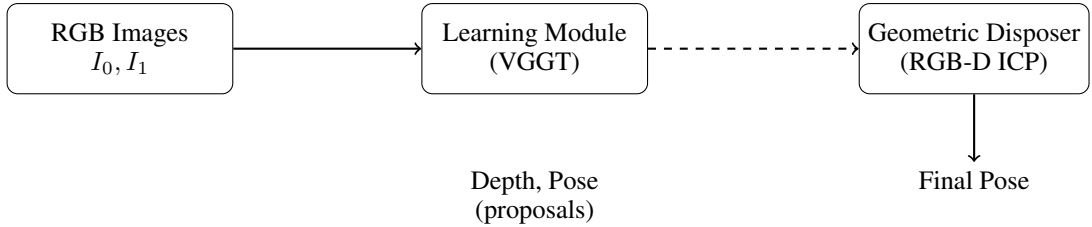
\begin{figure}[t]
\centering
\begin{tikzpicture}[
    box/.style={draw, rectangle, rounded corners, minimum width=3cm, minimum height=1.2cm, align=center},
    arrow/.style={->, thick},
    dashedarrow/.style={->, thick, dashed}
]

\node[box] (img) {RGB Images\\$I_0, I_1$};
\node[box, right=2.5cm of img] (learn) {Learning Module\\(VGGT)};
\node[box, right=2.8cm of learn] (geo) {Geometric Disposer\\(RGB-D ICP)};

\node[below=0.9cm of learn, align=center] (proposal) {Depth, Pose\\(proposals)};
\node[below=0.9cm of geo, align=center] (final) {Final Pose};

\draw[arrow] (img) -- (learn);
\draw[dashedarrow] (learn) -- (geo);
\draw[arrow] (geo) -- (final);

\end{tikzpicture}
\caption{Conceptual illustration of the proposed framework. Learning modules propose geometric observations, while pose estimation is validated and decided by an explicit geometric optimization backend. Dashed arrows indicate learning-based proposals without decision authority.}
\label{fig:framework}
\end{figure}

In this work, we instantiate this principle in the context of relative camera pose estimation on RGB-D sequences, as conceptually illustrated in Fig.~\ref{fig:framework}. We adopt VGGT as a representative learning model to produce geometric proposals, including dense depth and pose estimates, and pair it with a classical point-to-plane RGB-D ICP as the geometric disposer. Crucially, geometry is not treated as a minor refinement stage; instead, explicit geometric optimization is given full authority to validate and refine camera motion. Through controlled experiments on the TUM RGB-D benchmark, we systematically analyze when learning-based proposals help, when they fail, and how geometric disposal mediates their impact across different motion regimes and scene dynamics. Our goal is not to replace established geometric systems, but to clarify the role learning should play within geometry-centric pipelines.

%% file: related.tex
\label{sec:related_work}

\subsection{Learning-Based Pose Estimation}

Learning-based camera pose estimation has been extensively studied.
Early approaches regress camera poses directly from images using convolutional neural networks, exemplified by PoseNet~\cite{kendall2015posenet} and DeepVO~\cite{wang2017deepvo}.
More recent methods leverage large-scale training and stronger architectures to jointly infer pose, depth, or scene structure, achieving impressive performance under challenging visual conditions~\cite{teed2021droid,wang2025vggt}.

Despite these advances, most learning-based pose estimation methods treat network outputs as final estimates or strong priors.
Geometric consistency is often enforced implicitly through training objectives, such as photometric or reprojection losses~\cite{bian2019unsupervised}, rather than being explicitly verified at inference time.
As a result, erroneous predictions may propagate without a clear mechanism for rejection.

\subsection{Geometry-Centric Visual Odometry and SLAM}

Classical visual odometry and SLAM systems are built upon explicit geometric formulations, including multi-view geometry~\cite{hartley2003multiple}, rigid-body motion, and reprojection constraints.
Feature-based systems such as ORB-SLAM~\cite{campos2021orb} and dense RGB-D approaches such as KinectFusion~\cite{newcombe2011kinectfusion} enforce physical consistency through geometric optimization and have demonstrated strong robustness and interpretability.

However, geometry-centric methods are sensitive to initialization quality, data association errors, and sensor noise.
In particular, depth-based alignment methods relying on iterative closest point (ICP)~\cite{besl1992method} can fail under large inter-frame motion or poor initialization, motivating the integration of learning-based components to provide auxiliary information or improved priors.

\subsection{Hybrid Learning-Geometry Approaches}

A growing body of work combines learning-based perception with geometric optimization.
Common strategies include using learning to predict depth, optical flow, or compact scene representations, which are then consumed by a geometric backend for pose estimation~\cite{bloesch2018codeslam,czarnowski2020deepfactors}.
Other approaches integrate learning-based predictions directly into SLAM pipelines to improve robustness across sensing modalities~\cite{teed2021droid}.

While these hybrid systems benefit from both data-driven prediction and geometric constraints, learning outputs are typically integrated as soft priors or optimization aids.
Incorrect predictions are often absorbed by robust losses or heuristic filtering, without a clear separation between hypothesis generation and validation.
In contrast, our work emphasizes a strict separation between learning-based proposal generation and geometry-based decision making, where learning outputs are explicitly treated as hypotheses subject to geometric verification.

\subsection{Verification and Rejection in Estimation}

Robust estimation techniques for rejecting inconsistent measurements have long been studied in geometric vision.
Methods such as RANSAC~\cite{fischler1981random} and robust SLAM back-ends~\cite{sunderhauf2012towards,latif2014robust} address noise and outliers by filtering unreliable observations during optimization.

Our work differs by elevating rejection from a local robustness mechanism to a system-level principle.
Rather than filtering individual measurements, the proposed framework explicitly accepts or rejects entire learning-generated proposals based on geometric feasibility.
This reframes geometry as a decision boundary that governs the validity of learned hypotheses, providing a clear conceptual separation between learning-based proposal generation and geometry-based disposal.

%% file: problem.tex
This work studies visual pose estimation and mapping under a hybrid paradigm that separates \emph{proposal generation} from \emph{geometric verification}. We formalize the problem setting and clarify the roles of learning and geometry without committing to any specific algorithmic instantiation.

\subsection{Task Definition}

We consider the task of estimating camera poses and scene structure from a stream of visual observations.
Given a sequence of images $\{I_t\}_{t=1}^T$ (optionally accompanied by auxiliary sensory inputs), the goal is to estimate the corresponding camera poses $\{T_{w}^{t}\}$ and maintain a consistent geometric representation of the scene.

The system operates in an online, incremental manner, as opposed to an offline batch formulation.
This choice is motivated by practical deployment scenarios where observations arrive sequentially and timely pose estimates are required.
Moreover, an incremental setting exposes failure modes, such as drift accumulation and proposal bias, that are often obscured in offline pipelines relying on global initialization and batch optimization.

\subsection{Learning Proposals}

We define \emph{learning proposals} as hypotheses generated by data-driven models that provide coarse but structured predictions about the scene or camera motion.
These proposals may include, but are not limited to:
\begin{itemize}
    \item Relative or absolute camera pose hypotheses.
    \item Dense or sparse geometric predictions such as depth or correspondences.
    \item Other structural cues that suggest spatial relationships between views.
\end{itemize}
Importantly, learning proposals are treated as \emph{suggestions rather than commitments}. They encode statistical regularities learned from data and are not required to strictly satisfy geometric constraints.
This abstraction separates the formulation of proposals from their concrete realization, focusing on the role of learning-generated hypotheses rather than the specific mechanisms used to produce them. In this formulation, the learning component is accessed solely through its outputs, without assuming internal structure or enforcing explicit geometric constraints during its inference.

\subsection{Geometry Disposes}
The role of geometry is to \emph{dispose}, i.e., verify, refine, or reject learning proposals based on physical and geometric consistency.
Rather than generating hypotheses from scratch, the geometric module evaluates incoming proposals using principles such as:
\begin{itemize}
    \item multi-view consistency,
    \item reprojection agreement,
    \item physical feasibility under rigid-body motion,
    \item consistency with previously accepted estimates.
\end{itemize}

This design enforces hard structural constraints that are independent of data statistics.
Geometry therefore serves as a stabilizing mechanism that prevents error accumulation and mitigates failure modes of purely learning-based predictions.

\subsection{Evaluation Protocol}

To evaluate the effectiveness of the proposed paradigm, we consider metrics that measure both accuracy and robustness.
Typical evaluation criteria include trajectory accuracy (e.g., absolute and relative pose errors), tracking stability, and failure rates under challenging conditions.

We compare systems along three axes: learning-only approaches, geometry-only approaches, and hybrid systems that combine learning proposals with geometric disposal.
Detailed experimental settings and comparisons are deferred to Section~\ref{sec:experiments}.

%% file: framework.tex
\label{sec:pipeline}

This section presents a concrete instantiation of the proposed
\emph{learning proposes, geometry disposes} paradigm.
While the formulation in Section~2 is model-agnostic, we now describe the specific design choices adopted in our system.

\subsection{Learning Proposal Instantiation with VGGT}
\label{sec:learning_inst}

For the learning proposal module, we adopt VGGT as the proposal generator. We choose VGGT due to its ability to directly predict structured geometric quantities from monocular RGB inputs without relying on explicit feature matching or handcrafted geometric pipelines.
Given a pair of RGB images, VGGT produces coarse estimates of relative camera motion together with dense geometric predictions, including per-pixel depth and view-dependent correspondence cues.
These outputs serve as initial hypotheses about inter-frame geometry.

VGGT is used as-is, without finetuning or modification.
The system accesses the learning component solely through its outputs and does not impose explicit geometric constraints during inference.
As a result, the predicted poses and geometric quantities may be biased or inconsistent across views, especially in challenging scenes.

In our pipeline, although VGGT produces multiple geometric outputs, only the predicted depth maps are used in the geometric disposal stage.
Note that in our system all learning outputs are treated as proposals rather than final estimates, and geometric verification is performed exclusively using depth-based alignment.

\subsection{Geometry Disposal via ICP-based Verification}
\label{sec:geometry_disposal}

\begin{figure}[t]
    \centering
    \includegraphics[width=\linewidth]{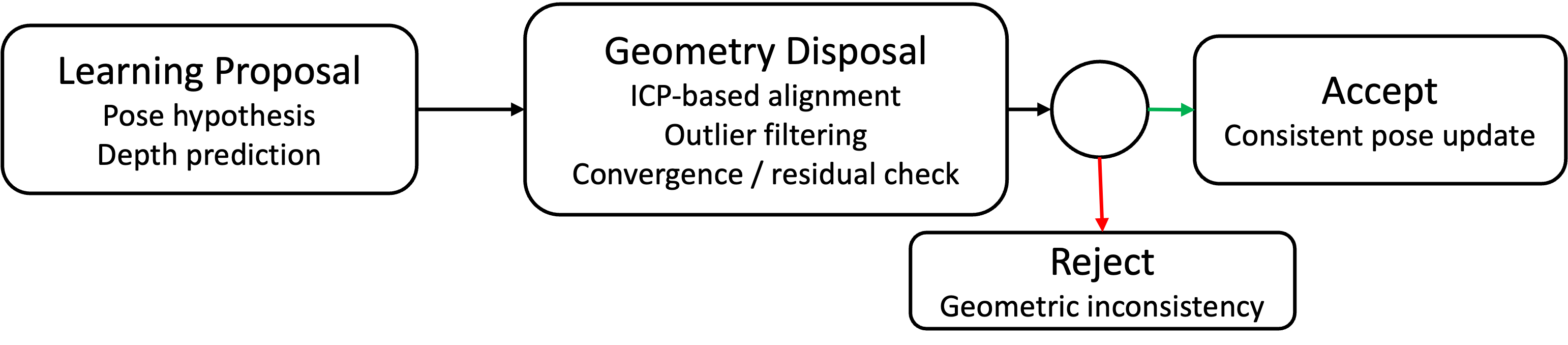}
    \caption{Geometry disposal semantics.
    A learning-generated proposal is evaluated by geometric consistency.
    Proposals that satisfy geometric constraints are accepted and integrated,
    while inconsistent proposals are explicitly rejected.}
    \label{fig:geometry_disposal}
\end{figure}

As illustrated in Fig.~\ref{fig:geometry_disposal}, the geometry disposal module serves as a decision boundary that evaluates learning-generated proposals and explicitly determines whether they are accepted or rejected based on geometric consistency.

To dispose learning proposals, we employ a geometry module based on iterative closest point (ICP) alignment.
ICP is chosen because it explicitly enforces geometric consistency through optimization failure and residual magnitude, naturally supporting proposal rejection in addition to refinement. In our implementation, this module is instantiated as a depth-based ICP operating on RGB-D observations.
Given a relative pose hypothesis $\hat{T}_{t}^{t+1}$ and geometric predictions from the learning stage, the goal is to verify and refine the proposed motion by enforcing geometric consistency.

Let $\mathcal{P}_t = \{\mathbf{p}_i\}$ denote a set of 3D points reconstructed from frame $t$, and $\mathcal{P}_{t+1} = \{\mathbf{q}_i\}$ the corresponding points from frame $t+1$.
ICP seeks an incremental rigid transformation $T \in SE(3)$ that minimizes the alignment error:
\begin{equation}
\label{eq:icp}
T^{*} = \arg\min_{T \in SE(3)} 
\sum_{i} \left\| \mathbf{q}_i - T \, \mathbf{p}_i \right\|^{2}.
\end{equation}

The ICP optimization is initialized using the pose proposal $\hat{T}_{t}^{t+1}$ produced by the learning module.
This initialization conditions the optimization on the learned hypothesis, while allowing hard geometric constraints to override it when inconsistencies arise.
During optimization, correspondences exhibiting large residuals are identified and filtered as outliers, preventing inconsistent geometric evidence from influencing pose refinement.

Beyond refinement, ICP functions as an explicit verification mechanism.
Proposals are discarded if the optimization fails to converge or if the final alignment residual exceeds a predefined threshold.
Only proposals that satisfy geometric consistency are accepted and propagated to subsequent stages.
By enforcing hard geometric constraints through Eq.~\eqref{eq:icp}, the disposal stage evaluates learning proposals against physical consistency and admits only those motions that satisfy geometric feasibility into the incremental pipeline.

\subsection{Incremental Pipeline Execution}
\label{sec:pipeline_execution}

The pipeline operates in an online, incremental manner.
At each time step, VGGT generates a learning proposal from the current observation pair.
This proposal is then refined and verified through ICP-based geometric disposal.
Accepted estimates are integrated into the current trajectory, while rejected proposals are discarded to avoid error accumulation.
This execution model exposes the interaction between learning bias and geometric constraints over time, which is difficult to observe in offline batch pipelines.

%% file: experiments.tex
\label{sec:experiments}

This section describes the experimental setup used to evaluate the proposed \emph{learning proposes, geometry disposes} paradigm.
We focus on controlled comparisons that isolate the effect of geometric disposal when learning-based proposals are introduced.

\subsection{Datasets}

We evaluate our system on the TUM RGB-D dataset, a widely used benchmark for RGB-D visual odometry and localization.
The dataset provides synchronized color and depth images together with accurate ground-truth camera trajectories, enabling quantitative evaluation of pose accuracy and robustness. All experiments are conducted in an incremental setting, where frames are processed sequentially without access to future observations. Although the TUM RGB-D dataset provides ground-truth depth measurements, we do not use them in the main experiments.
Instead, depth maps are predicted by the learning module for all evaluated methods.
This design ensures a consistent problem setting in which both pose and depth originate from learning-based proposals, allowing us to isolate the role of geometric disposal.
The effect of using ground-truth depth is analyzed separately as a sanity check.

\subsection{Compared Configurations}

To isolate the effect of geometric disposal and pose initialization, we consider the following baselines:
\begin{itemize}
    \item \textbf{Learning-only}: camera motion is estimated directly from the pose predictions produced by the learning module, without any geometric verification.
    
    \item \textbf{Geometry-only (Identity Init)}: camera motion is estimated using depth-based ICP initialized with the identity pose.
    The depth maps are predicted by the learning module, but no learning-based pose proposals are used.
    
    \item \textbf{Learning proposes + Geometry disposes (Ours)}: camera motion is estimated using depth-based ICP initialized with the pose proposals from the learning module, followed by geometric disposal.
\end{itemize}

\subsection{Evaluation Metrics}

We evaluate camera motion estimation using relative pose metrics that do not depend on global scale.
Specifically, we report relative pose error (RPE) in terms of rotational and translational drift over fixed frame intervals.
Absolute trajectory error (ATE) is not reported, as our setup does not estimate or enforce global scale.
All metrics are computed by aligning relative motions only, without scale correction.

\subsection{Implementation Details}
The learning proposal module is instantiated using a pretrained model and is used without finetuning.
For geometric disposal, we employ a depth-based ICP formulation operating on predicted depth maps.
Outlier rejection and convergence checks are applied during optimization to enable proposal disposal.
Unless otherwise specified, all methods use identical preprocessing and evaluation protocols to ensure fair comparison.

For each sequence, all methods are initialized identically and evaluated under the same conditions.
Learning-only and geometry-only baselines are executed using their respective components in isolation, while the proposed method integrates both through the disposal mechanism.
Quantitative results are reported as averages over all evaluated sequences.

%% file: results.tex
\label{sec:results}

We present quantitative results evaluating the effect of pose initialization and geometric disposal under different motion magnitudes and scene dynamics.
All results are reported using relative pose error (RPE) metrics.

\subsection{Overall Quantitative Results}

We evaluate the system on sequences from the TUM RGB-D dataset under increasing frame strides.
Larger strides correspond to larger inter-frame motion and more challenging geometric alignment.

Across all evaluated settings, geometric disposal significantly reduces the sensitivity of the final estimates to pose initialization.
While different initializations lead to noticeably different errors before disposal, their final performance after geometric verification is highly similar. As shown in Table~\ref{tab:summary_rpe}, the final relative pose errors after geometric disposal are highly similar for identity and learning-based pose initialization across all evaluated strides and sequences.

\subsection{Effect of Pose Initialization}

We first analyze the effect of pose initialization by comparing identity initialization and learning-based pose proposals.
Before geometric disposal, learning-based initialization does not consistently outperform identity initialization and, in several cases, exhibits larger rotational errors, particularly at higher strides.

After applying geometric disposal, however, the performance gap between different initializations largely disappears.
For both static and dynamic sequences, the final relative pose errors obtained with identity and learning-based initialization are nearly identical across all evaluated strides.
This indicates that geometric disposal effectively corrects or rejects erroneous pose proposals, rendering the system robust to initialization quality.

\subsection{Effect of Motion Magnitude}

As the frame stride increases, relative pose errors increase for all variants, reflecting the increased difficulty of the alignment problem.
This trend is consistent across static scenes (e.g., \texttt{fr1\_xyz}) and dynamic scenes (e.g., \texttt{fr3\_walking\_xyz}).

Notably, while larger strides amplify the differences between initial pose estimates, geometric disposal continues to constrain the final errors to a similar range.
This suggests that the proposed paradigm scales gracefully with motion magnitude by relying on geometric consistency rather than initialization accuracy alone.

\subsection{Static and Dynamic Scenes}
In dynamic scenes containing moving objects, all methods exhibit increased errors compared to static scenes.
Nevertheless, the relative behavior remains consistent: final estimates after geometric disposal show limited dependence on the choice of pose initialization.

These results indicate that geometric disposal provides a stabilizing effect even in the presence of dynamic elements, where learning-based pose predictions alone are prone to bias.

\subsection{Summary}
Overall, the results demonstrate that while learning-based pose proposals may provide useful priors in some cases, geometric disposal plays a decisive role in determining the final estimation quality.
By enforcing geometric consistency, the proposed system achieves robust performance that is largely insensitive to the quality of the initial pose proposals.

\paragraph{Sanity Check with Ground-Truth Depth.}
As a sanity check, we evaluate geometric disposal using ground-truth depth from the TUM RGB-D dataset on the \texttt{fr1\_xyz} sequence.
With oracle depth, learning-based pose initialization becomes more beneficial at larger strides, while identity initialization degrades more noticeably.
Nevertheless, geometric disposal remains essential for stabilizing estimation under large inter-frame motion.
Detailed quantitative results are reported in Appendix~\ref{sec:appendix_gt_depth}.

\begin{table}[t]
\centering
\caption{Final relative rotation error (RPE) after geometric disposal on the TUM RGB-D dataset.
Results are reported as mean / 95th percentile (degrees).
Identity and learning-based pose initializations yield highly similar final performance across different motion magnitudes and scene dynamics.}
\label{tab:summary_rpe}
\begin{tabular}{l c cc cc}
\toprule
\textbf{Sequence} & \textbf{Stride} &
\multicolumn{2}{c}{\textbf{Identity Init}} &
\multicolumn{2}{c}{\textbf{Learning Init}} \\
& & Mean & P95 & Mean & P95 \\
\midrule
fr1\_xyz & 1  & 0.690 & 1.803 & 0.690 & 1.803 \\
         & 5  & 2.653 & 7.614 & 2.604 & 7.190 \\
         & 10 & 5.254 & 12.982 & 4.780 & 13.514 \\
         & 15 & 7.619 & 21.119 & 6.086 & 15.687 \\
\midrule
fr3\_walking & 1  & 0.723 & 1.592 & 0.726 & 1.607 \\
             & 5  & 2.454 & 5.142 & 2.536 & 5.159 \\
             & 10 & 3.824 & 7.410 & 4.266 & 9.095 \\
             & 15 & 4.856 & 9.314 & 5.808 & 10.876 \\
\bottomrule
\end{tabular}
\end{table}

%% file: discussion.tex
\label{sec:discussion}

\subsection{Initialization Is a Proposal, Not a Commitment}

A key observation from our experiments is that learning-based pose initialization does not consistently outperform identity initialization prior to geometric verification.
At first glance, this result may appear counterintuitive, as learned pose predictions are often expected to provide strong priors.

However, this behavior highlights an important distinction between \emph{proposals} and \emph{commitments}.
In our framework, pose estimates produced by the learning module are treated strictly as hypotheses.
Geometric disposal subsequently determines whether these hypotheses are consistent with physical constraints, regardless of their origin.
As a result, the quality of the final estimates is largely determined by geometric consistency rather than the accuracy of the initial pose proposals.

\subsection{Geometry as a Decision Boundary}

Our results demonstrate that geometric reasoning serves as an explicit decision boundary that governs the acceptance or rejection of learning-generated proposals.
After geometric disposal, the final relative pose errors become largely insensitive to the choice of pose initialization, even under large inter-frame motion and dynamic scene content.

This behavior suggests that geometry plays a decisive role not merely as an optimizer, but as a mechanism for enforcing physical plausibility.
By explicitly rejecting inconsistent proposals, the system prevents erroneous predictions from propagating through the incremental pipeline.
This contrasts with end-to-end learning-based approaches, where errors are often implicitly absorbed into latent representations without explicit validation.

\subsection{Implications for Learning-Based Pose Estimation}

The observation that learning-based pose initialization does not dominate final performance raises broader questions about the role of learning in geometric estimation pipelines.
Rather than aiming to produce highly accurate pose estimates in isolation, learning modules may be more effective when designed to generate diverse, informative proposals that can be evaluated by downstream geometric constraints.

From this perspective, the objective of learning shifts from precise estimation to hypothesis generation.
This shift aligns naturally with the proposed paradigm, where learning explores the solution space and geometry enforces consistency.

\subsection{Limitations and Future Directions}

Our study focuses on depth-based geometric disposal using ICP, which enforces local geometric consistency.
While this choice allows us to clearly isolate the disposal mechanism, it does not capture all sources of geometric information, such as long-term loop closures or global optimization.

Future work may extend the proposed paradigm by incorporating richer geometric constraints, multi-frame verification, or uncertainty-aware disposal mechanisms.
Additionally, exploring learning models that explicitly account for their own uncertainty may further improve proposal quality without compromising the role of geometry as the final arbiter.

\subsection{Summary}

This work argues for a clear conceptual separation between learning-based proposal generation and geometry-based decision making in spatial estimation systems.
Through systematic experiments, we show that while learning-based pose predictions may provide informative hypotheses, they do not reliably determine final estimation quality.
Instead, geometric disposal plays a dominant role by explicitly enforcing physical consistency and rejecting invalid proposals.

A central finding of our study is that, after geometric disposal, the final performance becomes largely insensitive to the choice of pose initialization.
This observation holds across different motion magnitudes, scene dynamics, and initialization strategies.
Such behavior highlights that learning outputs should be interpreted as proposals rather than commitments, and that robustness emerges primarily from geometric validation rather than from improved initial predictions alone.

These results suggest a shift in how learning components should be designed and evaluated in geometry-centric systems.
Rather than optimizing for standalone accuracy, learning modules may be more effective when they generate diverse and informative hypotheses that facilitate downstream geometric verification.
In this sense, learning explores the solution space, while geometry determines feasibility.

Overall, our findings reinforce the proposed \emph{learning proposes, geometry disposes} paradigm.
By assigning learning and geometry distinct and complementary roles, the framework provides a principled pathway toward building spatial estimation systems that are both data-adaptive and physically grounded.

%% file: appendix.tex

\section{Additional Translational Error Statistics}
\label{sec:appendix_translation}
\begin{table}[h]
\centering
\caption{Final relative translational error (RPE) after geometric disposal on the TUM RGB-D dataset.
Results are reported as mean / 95th percentile.
Translational errors show minimal variation across different pose initialization strategies.}
\label{tab:appendix_translation}
\begin{tabular}{l c cc cc}
\toprule
\textbf{Sequence} & \textbf{Stride} &
\multicolumn{2}{c}{\textbf{Identity Init}} &
\multicolumn{2}{c}{\textbf{Learning Init}} \\
& & Mean & P95 & Mean & P95 \\
\midrule
fr1\_xyz & 1  & 0.023 & 0.048 & 0.023 & 0.048 \\
         & 5  & 0.104 & 0.204 & 0.102 & 0.204 \\
         & 10 & 0.203 & 0.429 & 0.193 & 0.386 \\
         & 15 & 0.265 & 0.616 & 0.259 & 0.548 \\
\midrule
fr3\_walking & 1  & 0.030 & 0.064 & 0.030 & 0.064 \\
             & 5  & 0.107 & 0.221 & 0.108 & 0.230 \\
             & 10 & 0.193 & 0.399 & 0.194 & 0.396 \\
             & 15 & 0.263 & 0.525 & 0.273 & 0.588 \\
\bottomrule
\end{tabular}
\end{table}
As shown in Table~\ref{tab:appendix_translation}, translational error statistics after geometric disposal are highly similar for identity and learning-based pose initialization across all evaluated settings.
Compared to rotational errors, translational errors exhibit limited sensitivity to initialization quality and are largely dominated by motion magnitude and depth scale.
For clarity, we therefore focus on rotational errors in the main paper.

\section{Sanity Check with Ground-Truth Depth}
\label{sec:appendix_gt_depth}

The results in Table~\ref{tab:appendix_gt_depth} report geometric disposal performance using ground-truth depth.
Compared to learned depth, oracle depth reduces alignment ambiguity and exposes the effect of pose initialization more clearly, particularly at larger strides.
This experiment provides an upper-bound reference and is not intended as a direct comparison with the main experimental setting using learned depth.

\begin{table}[h]
\centering
\caption{Final relative rotation error (RPE) after geometric disposal using ground-truth depth from the TUM RGB-D dataset on the \texttt{fr1\_xyz} sequence.
Results are reported as mean / 95th percentile (degrees).
This experiment serves as a sanity check to contextualize performance with oracle depth.}
\label{tab:appendix_gt_depth}
\begin{tabular}{c cc cc}
\toprule
\textbf{Stride} &
\multicolumn{2}{c}{\textbf{Identity Init}} &
\multicolumn{2}{c}{\textbf{Learning Init}} \\
 & Mean & P95 & Mean & P95 \\
\midrule
1  & 0.688 & 1.803 & 0.691 & 1.812 \\
5  & 2.805 & 7.138 & 2.747 & 7.139 \\
10 & 5.390 & 13.422 & 4.892 & 12.504 \\
15 & 9.100 & 24.949 & 6.279 & 14.782 \\
\bottomrule
\end{tabular}
\end{table}

%% file: iclr2026_conference.bib
@inproceedings{wang2025vggt,
  title={Vggt: Visual geometry grounded transformer},
  author={Wang, Jianyuan and Chen, Minghao and Karaev, Nikita and Vedaldi, Andrea and Rupprecht, Christian and Novotny, David},
  booktitle={Proceedings of the Computer Vision and Pattern Recognition Conference},
  pages={5294--5306},
  year={2025}
}

@InProceedings{sturm12iros,
	author = {J. Sturm and N. Engelhard and F. Endres and W. Burgard and D. Cremers},
	title = "A Benchmark for the Evaluation of RGB-D SLAM Systems",
	booktitle = "Proc. of the International Conference on Intelligent Robot Systems (IROS)",
	year = "2012",
	month= "Oct.",
}

@book{hartley2003multiple,
  title={Multiple view geometry in computer vision},
  author={Hartley, Richard and Zisserman, Andrew},
  year={2003},
  publisher={Cambridge university press}
}

@article{davison2007monoslam,
  title={MonoSLAM: Real-time single camera SLAM},
  author={Davison, Andrew J and Reid, Ian D and Molton, Nicholas D and Stasse, Olivier},
  journal={IEEE transactions on pattern analysis and machine intelligence},
  volume={29},
  number={6},
  pages={1052--1067},
  year={2007},
  publisher={IEEE}
}

@article{campos2021orb,
  title={Orb-slam3: An accurate open-source library for visual, visual--inertial, and multimap slam},
  author={Campos, Carlos and Elvira, Richard and Rodr{\'\i}guez, Juan J G{\'o}mez and Montiel, Jos{\'e} MM and Tard{\'o}s, Juan D},
  journal={IEEE transactions on robotics},
  volume={37},
  number={6},
  pages={1874--1890},
  year={2021},
  publisher={IEEE}
}

@inproceedings{triggs1999bundle,
  title={Bundle adjustment—a modern synthesis},
  author={Triggs, Bill and McLauchlan, Philip F and Hartley, Richard I and Fitzgibbon, Andrew W},
  booktitle={International workshop on vision algorithms},
  pages={298--372},
  year={1999},
  organization={Springer}
}

@inproceedings{besl1992method,
  title={Method for registration of 3-D shapes},
  author={Besl, Paul J and McKay, Neil D},
  booktitle={Sensor fusion IV: control paradigms and data structures},
  volume={1611},
  pages={586--606},
  year={1992},
  organization={Spie}
}

@inproceedings{klein2007parallel,
  title={Parallel tracking and mapping for small AR workspaces},
  author={Klein, Georg and Murray, David},
  booktitle={2007 6th IEEE and ACM international symposium on mixed and augmented reality},
  pages={225--234},
  year={2007},
  organization={IEEE}
}

@article{mur2015orb,
  title={ORB-SLAM: A versatile and accurate monocular SLAM system},
  author={Mur-Artal, Raul and Montiel, Jose Maria Martinez and Tardos, Juan D},
  journal={IEEE transactions on robotics},
  volume={31},
  number={5},
  pages={1147--1163},
  year={2015},
  publisher={IEEE}
}

@article{teed2021droid,
  title={Droid-slam: Deep visual slam for monocular, stereo, and rgb-d cameras},
  author={Teed, Zachary and Deng, Jia},
  journal={Advances in neural information processing systems},
  volume={34},
  pages={16558--16569},
  year={2021}
}

@article{teed2018deepv2d,
  title={Deepv2d: Video to depth with differentiable structure from motion},
  author={Teed, Zachary and Deng, Jia},
  journal={arXiv preprint arXiv:1812.04605},
  year={2018}
}

@inproceedings{wang2021tartanvo,
  title={Tartanvo: A generalizable learning-based vo},
  author={Wang, Wenshan and Hu, Yaoyu and Scherer, Sebastian},
  booktitle={Conference on Robot Learning},
  pages={1761--1772},
  year={2021},
  organization={PMLR}
}

@inproceedings{wang2017deepvo,
  title={Deepvo: Towards end-to-end visual odometry with deep recurrent convolutional neural networks},
  author={Wang, Sen and Clark, Ronald and Wen, Hongkai and Trigoni, Niki},
  booktitle={2017 IEEE international conference on robotics and automation (ICRA)},
  pages={2043--2050},
  year={2017},
  organization={IEEE}
}

@article{bian2019unsupervised,
  title={Unsupervised scale-consistent depth and ego-motion learning from monocular video},
  author={Bian, Jiawang and Li, Zhichao and Wang, Naiyan and Zhan, Huangying and Shen, Chunhua and Cheng, Ming-Ming and Reid, Ian},
  journal={Advances in neural information processing systems},
  volume={32},
  year={2019}
}

@inproceedings{newcombe2011kinectfusion,
  title={Kinectfusion: Real-time dense surface mapping and tracking},
  author={Newcombe, Richard A and Izadi, Shahram and Hilliges, Otmar and Molyneaux, David and Kim, David and Davison, Andrew J and Kohi, Pushmeet and Shotton, Jamie and Hodges, Steve and Fitzgibbon, Andrew},
  booktitle={2011 10th IEEE international symposium on mixed and augmented reality},
  pages={127--136},
  year={2011},
  organization={Ieee}
}

@article{czarnowski2020deepfactors,
  title={Deepfactors: Real-time probabilistic dense monocular slam},
  author={Czarnowski, Jan and Laidlow, Tristan and Clark, Ronald and Davison, Andrew J},
  journal={IEEE Robotics and Automation Letters},
  volume={5},
  number={2},
  pages={721--728},
  year={2020},
  publisher={IEEE}
}

@inproceedings{bloesch2018codeslam,
  title={Codeslam—learning a compact, optimisable representation for dense visual slam},
  author={Bloesch, Michael and Czarnowski, Jan and Clark, Ronald and Leutenegger, Stefan and Davison, Andrew J},
  booktitle={Proceedings of the IEEE conference on computer vision and pattern recognition},
  pages={2560--2568},
  year={2018}
}

@inproceedings{sunderhauf2012towards,
  title={Towards a robust back-end for pose graph SLAM},
  author={S{\"u}nderhauf, Niko and Protzel, Peter},
  booktitle={2012 IEEE international conference on robotics and automation},
  pages={1254--1261},
  year={2012},
  organization={IEEE}
}

@inproceedings{latif2014robust,
  title={Robust graph SLAM back-ends: A comparative analysis},
  author={Latif, Yasir and Cadena, C{\'e}sar and Neira, Jos{\'e}},
  booktitle={2014 IEEE/RSJ International Conference on Intelligent Robots and Systems},
  pages={2683--2690},
  year={2014},
  organization={IEEE}
}

@article{fischler1981random,
  title={Random sample consensus: a paradigm for model fitting with applications to image analysis and automated cartography},
  author={Fischler, Martin A and Bolles, Robert C},
  journal={Communications of the ACM},
  volume={24},
  number={6},
  pages={381--395},
  year={1981},
  publisher={ACM New York, NY, USA}
}

@inproceedings{kendall2015posenet,
  title={Posenet: A convolutional network for real-time 6-dof camera relocalization},
  author={Kendall, Alex and Grimes, Matthew and Cipolla, Roberto},
  booktitle={Proceedings of the IEEE international conference on computer vision},
  pages={2938--2946},
  year={2015}
}
